\documentclass[acmtog]{acmart}

\renewcommand\footnotetextcopyrightpermission[1]{} 
\AtBeginDocument{%
  \providecommand\BibTeX{{%
    \normalfont B\kern-0.5em{\scshape i\kern-0.25em b}\kern-0.8em\TeX}}}

\copyrightyear{2023}
\acmYear{2023}
\setcopyright{acmlicensed}\acmConference[SIGGRAPH '23 Conference Proceedings]{Special Interest Group on Computer Graphics and Interactive Techniques Conference Conference Proceedings}{August 6--10, 2023}{Los Angeles, CA, USA}
\acmBooktitle{Special Interest Group on Computer Graphics and Interactive Techniques Conference Conference Proceedings (SIGGRAPH '23 Conference Proceedings), August 6--10, 2023, Los Angeles, CA, USA}
\acmPrice{15.00}
\acmDOI{10.1145/3588432.3591555}
\acmISBN{979-8-4007-0159-7/23/08}

\begin{document}

\title{NOFA: NeRF-based One-shot Facial Avatar Reconstruction}

\author{Wangbo Yu}
\affiliation{
    \institution{Tencent AI Lab}
    \country{China}
    }

\author{Yanbo Fan}
\authornote{Yanbo Fan, Yong Zhang and Xuan Wang are the corresponding authors.}
\affiliation{
    \institution{Tencent AI Lab}
    \country{China}
    }
    
\author{Yong Zhang}
\authornotemark[1]
\affiliation{
    \institution{Tencent AI Lab}
    \country{China}
    }
    
\author{Xuan Wang}
\authornotemark[1]
\affiliation{
    \institution{Ant Group}
    \country{China}
    }
    
\author{Fei Yin}
\affiliation{
    \institution{Tsinghua University}
    \country{China}
    }
    
\author{Yunpeng Bai}
\affiliation{
    \institution{Tsinghua University}
    \country{China}
    }
    
\author{Yan-Pei Cao}
\affiliation{
    \institution{Tencent AI Lab}
    \country{China}
    }
    
\author{Ying Shan}
\affiliation{
    \institution{Tencent AI Lab}
    \country{China}
    }
    
\author{Yang Wu}
\affiliation{
    \institution{Tencent AI Lab}
    \country{China}
    }

\author{Zhongqian Sun}
\affiliation{
    \institution{Tencent AI Lab}
    \country{China}
    }
    
\author{Baoyuan Wu}
\affiliation{
    \institution{The Chinese University of Hong Kong, Shenzhen}
    \country{China}
    }

\renewcommand{\shortauthors}{Yu et al.}

\begin{abstract}
    
3D facial avatar reconstruction has been a significant research topic in computer graphics and computer vision, where photo-realistic rendering and flexible controls over poses and expressions are necessary for many related applications.
Recently, its performance has been greatly improved with the development of neural radiance fields (NeRF).
However, most existing NeRF-based facial avatars focus on subject-specific reconstruction and reenactment, requiring multi-shot images containing different views of the specific subject for training, and the learned model cannot generalize to new identities, limiting its further applications.
In this work, we propose a one-shot 3D facial avatar reconstruction framework that only requires a single source image to reconstruct a high-fidelity 3D facial avatar.
For the challenges of lacking generalization ability and missing multi-view information, we leverage the generative prior of 3D GAN and develop an efficient encoder-decoder network to reconstruct the canonical neural volume of the source image, and further propose a compensation network to complement facial details.
To enable fine-grained control over facial dynamics, we propose a deformation field to warp the canonical volume into driven expressions.
Through extensive experimental comparisons, we achieve superior synthesis results compared to several state-of-the-art methods.

\end{abstract}


\begin{teaserfigure}
\centering
  \includegraphics[width=1.0\textwidth]{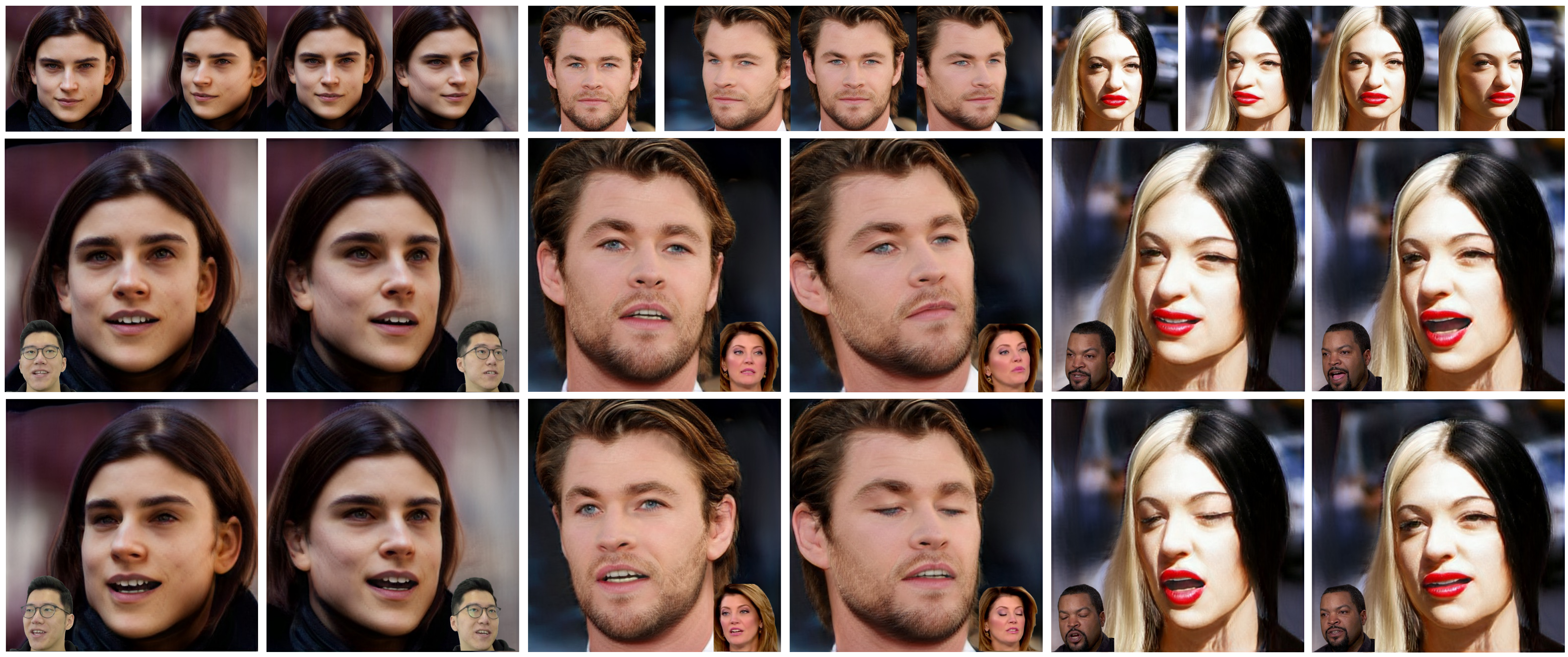}
  \caption{Our method enables high-fidelity facial avatar reconstruction and reenactment given a single input image. The first row shows the input image and novel view synthesis results, the following rows show the facial reenactment results, where the facial motion of the avatars are controlled by the driving faces.}
  \label{fig:teaser}
\end{teaserfigure}

\maketitle

\section{Introduction}
\label{sec:introduction}

Facial avatar reconstruction has been an important research topic in the field of computer graphics and computer vision, due to its phenomenal applications in virtual reality (VR), augmented reality (AR), movie industry, and teleconferencing.  
High-fidelity face reconstruction and fine-grained face reenactment are foundations for those applications.  

To animate facial images, several 2D approaches have been proposed by utilizing flow-based warping in image or feature spaces to 
transfer motion, and encoder-decoder networks to synthesize appearance \cite{yin2022styleheat,siarohin2019first,wang2021one, wang2022latent, siarohin2019animating, drobyshev2022megaportraits,cheng2022videoretalking}. 
By training on large-scale face video datasets \cite{chung2018voxceleb2,wang2021one,zhu2022celebv} with a large number of identities, these methods are generalizable to new identities and can produce vivid reenactment results given just a single facial image of the source identity.
However, they have no constraints on the underlying 3D facial geometry and can hardly generate multi-view consistent images. 
Besides, they suffer from artifacts under large driven poses or expressions. 
Meanwhile, conventional parametric face model \cite{blanz1999morphable, paysan20093d} -based methods \cite{deng2019accurate,feng2021learning,garrido2016reconstruction,sanyal2019learning, tewari2019fml,xing2023codetalker} model 3D faces with template mesh and 3DMM parameters \cite{blanz1999morphable}.
They support the flexible controls over poses and expressions.
However, these mesh-based methods are memory-inefficient and less effective in modeling the non-face region, such as teeth, hair, and accessories. 
The accuracy of reconstruction and animation is also limited by the number of blendshapes or templates.

Recently, the photo-realistic and multi-view consistent rendering ability of Neural Radiance Fields (NeRF) \cite{mildenhall2020nerf} has sparked several works for NeRF-based facial avatar reconstruction \cite{athar2022rignerf,grassal2022neural,gafni2021dynamic}. 
They perform facial reenactment by learning deformation field or rendering function conditioned on control signals, where the pose and expression coefficients of 3DMMs are most commonly used.
There exist two limitations of those methods. 
First, they require a large number of images containing different poses and expressions of the target face for training, which are not always available in real scenarios. 
Second, they are subject-dependent that can only be used to generate images of the training identity, \textit{i.e.,} they are not generalizable to new identities.
The lack of generalization ability and the demand on extensive multi-shot training data limit their further applications. 

To address the aforementioned limitations of the existing methods, we aim to propose a generalizable NeRF-based one-shot facial avatar reconstruction method, which can be applicable to any new identity given only a single facial image.
There are three main challenges for the considered objective: \textbf{1)} due to the complex facial dynamics and missing 3D information, it is challenging to learn a faithful reconstruction from a single input image, \textbf{2)} it is difficult to endow generalization ability for personalized NeRF, and \textbf{3)} fine-grained control over facial expression remains an challenging task in NeRF, especially for the one-shot situation.
To tackle the challenges, we take advantage of the generative prior of a NeRF-based 3D GAN \cite{eg3d}. 
The latent space of 3D GAN encodes rich 3D-consistent generative prior, which helps synthesize neural volumes of diverse human faces.
To align the unconditional latent space with real images, we develop a model with the encoder-decoder framework, and train it on a large-scale video dataset \cite{zhu2022celebv} in an end-to-end manner.
Specifically, the encoder projects the input image to the latent space while the decoder maps the latent code to a neural volume, which is then used for image synthesis via neural rendering.
Different from GAN inversion that learns a latent code that can accurately reconstruct the input image, here we aim to project the input image into a shared canonical space with an aligned expression, which is crucial for modeling facial dynamics.
Due to the inevitable information loss of the encoder, the reconstructed volume lacks image-specific details, leading to poor identity preservation.
To supplement identity information leakage, we design a compensation network to learn a compensatory neural volume based on the input image and the intermediate feature from the decoder. 

For the sake of fine-grained face reenactment, we exploit a deformation field to model facial dynamics, which learns a conditional mapping to deform each sampled point in the target space into the canonical space according to the driven signals.
We extend the personalized deformation field in existing works to a generalized deformation field to handle different test identities and driven signals, by conditioning it on both identity and expression coefficients of 3DMMs and training it on the large-scale video dataset \cite{zhu2022celebv}.
Thanks to the rich training expressions offered by the dataset, our deformation field can better model large and extreme motions compared with the personalized methods.
We conduct extensive experiments and compare to both 2D and NeRF-based face reconstruction and reenactment methods, and demonstrate our superior performance in novel view synthesis and reenactment tasks.

Our main contributions are in three-fold: \textbf{1)} we propose a NeRF-based one-shot facial avatar reconstruction method that supports high-fidelity 3D face reconstruction and vivid face reenactment from a single face image.
\textbf{2)} Once trained, our method can be generalized to new identities, which is more efficient and practical than personalized methods. 
\textbf{3)} We compare our method with several state-of-the-art methods and obtain superior synthesis performance.
\section{Related Work}
\label{sec:relatedwork}

\subsection{Neural Scene Representation} 
Neural radiance fields (NeRF) \cite{mildenhall2020nerf} represents 3D scenes using MLP-based implicit function and achieve compelling rendering quality in 3D reconstruction tasks. 
Benefiting from its inherently differentiable rendering process, NeRF can be trained simply using multi-view images and their corresponding camera labels, and has been widely used in the field of 3D modeling and novel view synthesis. 
However, the conventional NeRF cannot handle dynamic subjects.
Several approaches have been devoted to work around this limitation \cite{athar2022rignerf,grassal2022neural,gafni2021dynamic,guo2021ad,wang2021learning, raj2021pixel, su2021anerf, xian2021space,tretschk2021non, park2021nerfies}.
The solutions can be roughly categorized into two categories: a train of thought is to condition the radiance field on control signals, which will change the density and color of the observed points. 
Another train of thought is to additionally learn a deformation field that accepts control signals and coordinates as input and predicts coordinate offsets from the deformed space into canonical space
For NeRF-based facial avatar synthesis \cite{athar2022rignerf,grassal2022neural,gafni2021dynamic,guo2021ad,zhang2022fdnerf,gao2022reconstructing,zheng2022avatar},the pose and expression coefficients of 3D Morphable Face Models (3DMMs) \cite{blanz1999morphable} are employed as control signals to model facial deformations. 
These works study subject-dependent reconstruction and cannot generalize to different identities.
What's more, a large set of facial images of the given identity are needed for training.
Differently, we study the subject-agnostic problem,  
where only a single portrait image is given for reconstruction, and propose a generalizable model that can handle different testing faces.

\begin{figure*}[htbp]
  \centering
  \includegraphics[width=0.85\textwidth]{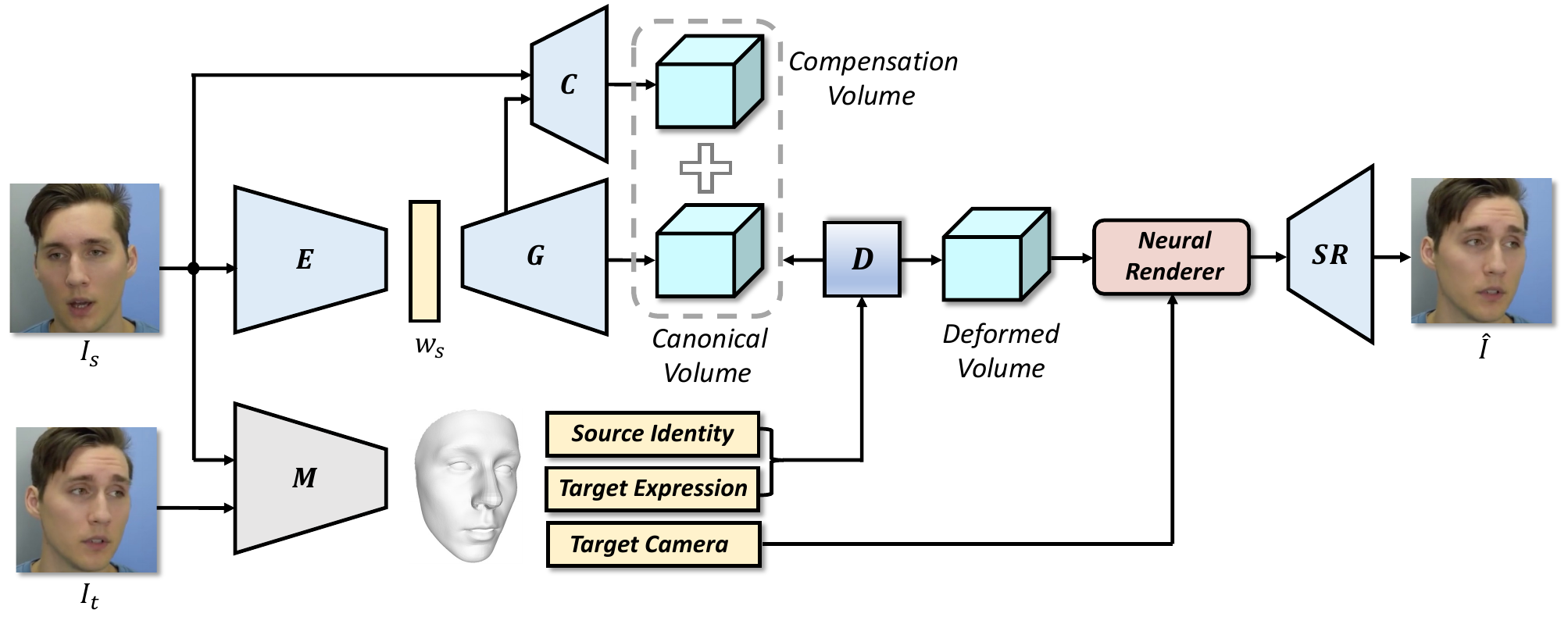}
  \vspace{2pt}
  \caption{Pipeline of NOFA. Given a source image $I_s$, an encoder $E$ is adopted to embed the image into the latent space of volume generator $G$, which will produce the canonical volume $V_c$ with an aligned expression while preserve the identity of $I_s$. The compensation network $C$ is used to supplement image-specific details for $V_c$. To achieve explicit motion control, we employ pretrained $M$ to extract 3DMM parameters from target image $I_t$ and source image $I_s$, and use the combination of source identity and target expression parameters as control signal of the deformation field $D$, to deform the canonical volume with source identity into target expression. Finally, a hybrid neural renderer consists of volume rendering and super-resolution ($SR$) modules is adopted to render the final output $\hat{I}$ given target camera parameters.}
\label{pipeline}
\end{figure*}
\subsection{3D-aware Generative Networks}
Inspired by the breakthroughs achieved by 2D Generative Adversarial Networks (GANs) \cite{karras2019style,karras2020analyzing,karras2021alias}, recent researches \cite{chan2021pi, schwarz2020graf} have extended 2D image generation into 3D settings by combining GANs with the Implicit Neural Representations (INRs).  
These unconditional 3D GANs can generate photo-realistic rendering and enable controls over views. 
However, they do not support fine-grained and explicit expression controls.
Recently \cite{wang2022morf} proposed a generative NeRF that overfits multiple identities at the same time, by learning subject-specific identity codes as the condition of NeRF MLPs. This method can be used for one-shot head avatar synthesis by fine-tuning the latent code and MLP parameters on a single source image. However, its training data are captured in studio conditions, and the one-shot synthesis results are of low quality due to its sparse latent space (only 15 identities are encoded). 
Some concurrent works 
\cite{bergman2022gnarf,wu2022anifacegan,sun2022next3d,tang2022explicitly} further utilize the parameters of 3D Morphable Face Models (3DMMs) to explicitly control expression of the rendered faces.
Yet, they are designed for unconditionally generating fake images and cannot be directly used in real applications.
Besides, it is challenging for them to generate fine-grained motion because they are trained on discrete face image datasets. 
Different from these methods, we make use of the prior of 3D GAN and jointly train an encoder-decoder network on large-scale video datasets, achieving real image 3D reconstruction and vivid motion reenactment. 

\subsection{GAN Inversion}
GAN inversion techniques act as a bridge for bring GANs to real world applications such as image editing and reenactment. Existing GAN inversion approaches can be roughly divided into three categories: the optimization-based methods~\cite{abdal2019image2stylegan,abdal2020image2stylegan++}, which optimize the latent codes by minimizing the distance between the ground truth image and the generated one, achieving promising reconstruction results yet limited by its low efficiency. The learning-based methods utilize an encoder network to directly encode the input images into latent codes ~\cite{richardson2021encoding} ~\cite{alaluf2021restyle}, equipping with high efficiency and generalization ability while the reconstruction results often lacks fine details due to the information loss in the encoding process.  
The hybrid GAN inversion approaches utilize a learned encoder to predict an initial latent code and further refine it in the optimization process~\cite{zhu2020indomain} ~\cite{yin20223d}, the generator parameters are also optimized in  ~\cite{roich2021pivotal} to achieve better results. 
Despite their success in real image editing, GAN inversion is usually applied for editing global facial attributes such as age, makeup and gender. It is non-trivial to be used for fine-grained face reenactment.

\subsection{One-shot Talking Head Synthesis} 
One-shot talking head synthesis aims to generate talking face videos from a given source image and a driving video.
The generated videos should maintain the facial characteristics of the source image and the facial movements in the driving video.
A large amount of works study these in the 2D image or feature space, where the key idea is to learn two separated networks to control motion and model appearance. For example, the works of \cite{siarohin2019first,siarohin2019animating} predict warping flow from key-points to warp features of source images into target motion. 
The work of \cite{ren2021pirenderer} uses 3DMM parameters to modulate flow generator and a refine network to supplement fine details. \cite{yin2022styleheat} further leverages the prior of StyleGAN \cite{karras2019style} to enhance appearance. These methods are trained on the large-scale face video datasets \cite{chung2018voxceleb2,wang2021one,zhu2022celebv}  containing rich identities and expressions, thus can be generalized to unseen motion and identity given just a single input image. However, they cannot handle large pose changes due to the artifacts brought by feature warping. Some methods \cite{drobyshev2022megaportraits, wang2021one} have devoted to address this problem by introducing 3D CNNs to produce 3D feature representation of the input image and apply 3D feature warping.
Nevertheless, the learned representation doesn't model the underlying 3D facial geometry and the warping process lack explicit 3D constraints, causing poor multi-view consistency and can hardly be used in novel view synthesis.
\section{Method}
\label{sec:method}

We propose NOFA, a NeRF-based one-shot facial avatar reconstruction framework, which will be described in detail in this section. We first introduce the networks employed for image to volume synthesis in Sec.~\ref{sec:3.1}. Then, we present the details of the 3DMM-guided deformation field for dynamic modeling in Sec.~\ref{sec:3.2}. Finally, we provide the loss functions used in the training stage as well as explaining the training strategy in Sec.~\ref{sec:3.3}.

 \subsection{Volume Synthesis with Generative Prior}
\label{sec:3.1}

In order to reconstruct high-fidelity facial avatars, the first and most important step is to build 3D representation of the given subject. 
As shown in Fig.~\ref{pipeline}, we leverage the generator of a pretrained state-of-the-art 3D GAN \cite{eg3d} to synthesize 3D representation and render images, which consists of a tri-plane generator $G$, a volumetric rendering module, and a super resolution module.  
$G$ uses a StyleGAN2 \cite{karras2020analyzing} backbone to synthesize features of size $96\times256\times256$, which will be reshaped into
a tri-plane of size $3\times32\times256\times256$. Then,  features of each position in the tri-plane could be efficiently queried by coordinates and decoded to neural volumes for volumetric rendering, producing high-fidelity and view-consistent images.

We draw ideas from general GAN inversion approaches and achieve image to volume synthesis by exploiting 
learning-based GAN inversion. 
Specifically, given a source facial image $I_s$, we adopt a standard e4e \cite{tov2021designing} encoder to project the image to the latent space of $G$, \textit{i.e.,} embedding $I_s$ into a set of latent codes $w_s$ 
that will modulate features in the tri-plane generator $G$ for volume generation:
\begin{equation}
    V_c = G(E(I_s) + \overline{w}),
\end{equation}
where $\overline{w}$ is the average latent code of $G$, and $V_c$ denotes the output neural volume. 
Different from conventional inversion approaches that  faithfully reconstruct the input image, the generated volume $V_c$ is defined in the canonical space, \textit{i.e.}, with an aligned expression instead of preserving the original expression of $I_s$. This is crucial for the following deformation process, where we gain explicit expression control by exploiting backward deformation. In our implementation, the canonical space is naturally learned by jointly training the whole framework on videos in an end-to-end manner, without explicit inner supervision. 

As conjectured in \cite{blau2019rethinking} that the low-rate latent codes produced by the encoder are insufficient for high-fidelity reconstruction, we thus additionally learn a compensation network $C$ to compensate the information loss caused by $E$.

\begin{figure*}[t]
  \centering
  \includegraphics[width=.8\linewidth]{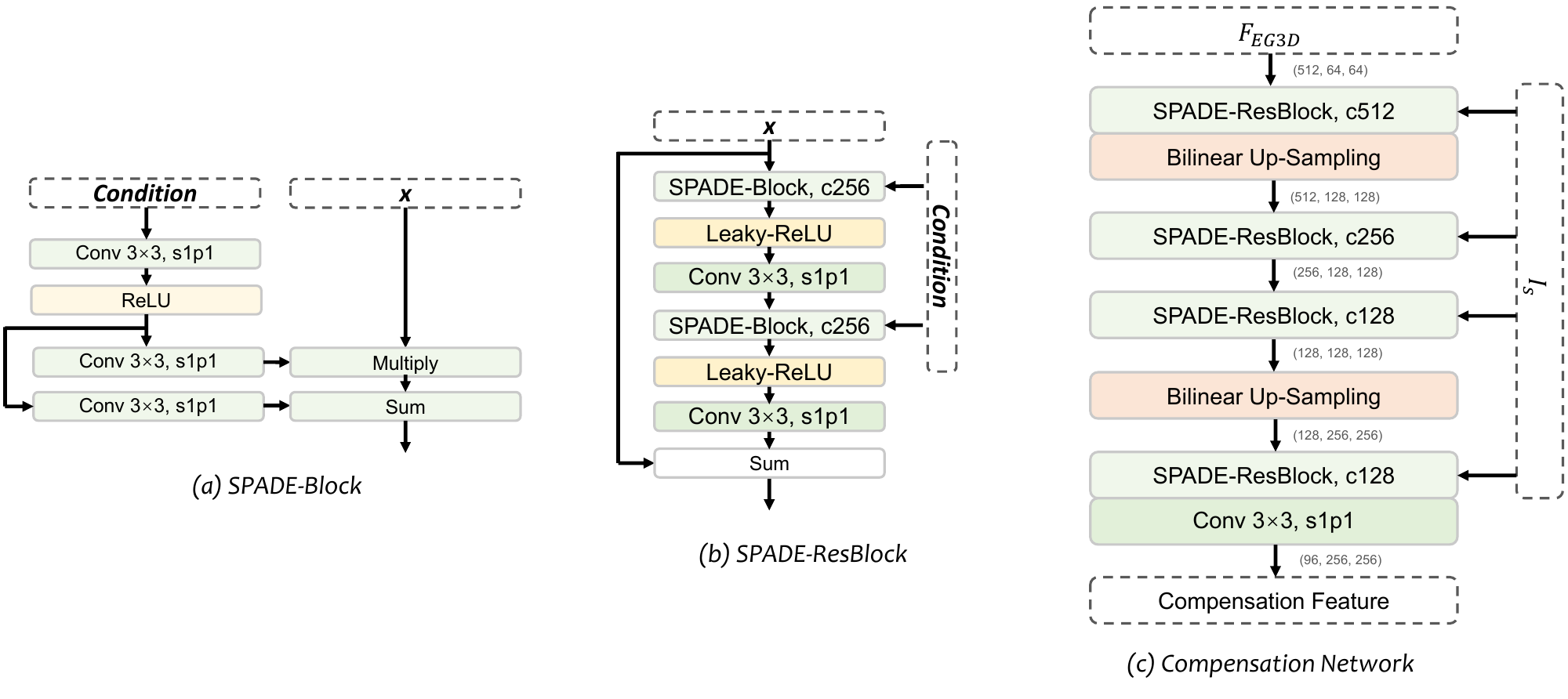}
  \caption{Architecture of the compensation network, which comprises multiple blocks of spatially-adaptive de-normalization (SPADE) and convolution. It takes the intermediate feature of tri-plane generator $G$ and source image $I_s$ as input and outputs the compensation volume for supplementing identity and texture information.
In SPADE, the intermediate features are progressively modulated with a set of scale and shift parameters predicted from $I_s$.
We discard the batch normalization (BN) layers in SPADE to better preserve the information of the intermediate features. }
  \label{fig:spade}
\end{figure*}

Fig.~\ref{fig:spade} (c) shows the architecture of the compensation network. It contains several spatially-adaptive de-normalization (SPADE) res-blocks \cite{park2019semantic} and convolution,  which takes the $64 \times 64 \times 512$ feature of the tri-plane generator $G$ as input and progressively modulate and upscale the feature with source image $I_s$. 
The details of SPADE-ResBlock are given in Fig. ~\ref{fig:spade} (a) and Fig.~\ref{fig:spade} (b). We discard the batch normalization (BN) layers in SPADE to better preserve the tri-plane features. 
The final output compensation volume is of the same size with the output of $G$. They are summed together to produce the neural volume with better identity and appearance preservation. 

\subsection{Dynamic Modeling with 3DMM Guidance}
\label{sec:3.2}
In order to achieve explicit motion control on the reconstructed neural volume, we exploit a deformation field $D$ to model facial dynamics and employ the semantic parameters of a 3DMM face model \cite{paysan20093d} as control signals. In 3DMM, the face shape is defined as:
\begin{equation}
    \mathbf{S} = \bar{\mathbf{S}} + \alpha\mathbf{{B}_{id}} + \beta\mathbf{{B}_{exp}},
\end{equation}
where $\bar{\mathbf{S}}$ represents the average face shape, $\mathbf{{B}_{id}}$ and $\mathbf{{B}_{exp}}$ are the identity and expression basis computed by PCA~\cite{jolliffe2016principal}. 
We adopt the coefficients $\alpha$ and $\beta$ as control signals, which are semantically meaningful and enable fine-detailed expression control. 

During training, we employ an off-the-shelf 3D face reconstruction model \cite{deng2019accurate} to estimate source identity parameter $\alpha_s$, target expression parameter $\beta_t$, and target camera parameter $c_t$ from target image $I_t$ and source image $I_s$, and use the combination of $\alpha_s$ and $\beta_t$ as the input control signal of the deformation field. 
Then, a hybrid neural renderer consisting of volume rendering and super-resolution ($SR$) is adopted to render the final output $\hat{I}$ given target camera parameters $c_t$ that model head rotation and translation.

\begin{figure}[t]
  \centering
  \includegraphics[width=\linewidth]{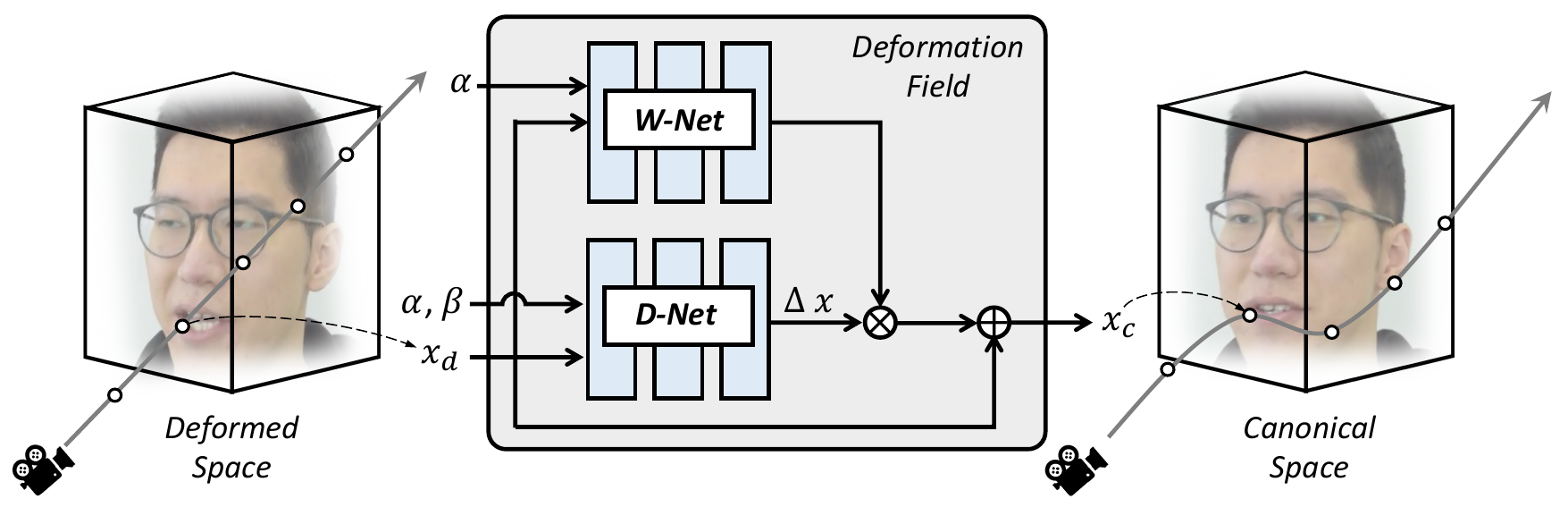}
  \caption{Illustration of the deformation field. It consists of a deformation network (D-Net) and a weighting network (W-Net). D-Net regress the coordinates offsets from the deformed space to the canonical space. W-Net predicts the per-location weight scalars to multiply with the offsets. With the weighted offsets, we can query volume features defined in the canonical space for volumetric rendering.}
  \Description{}
  \label{fig:deformtion}
\end{figure}

In particular, the deformation field models the backward deformation that deforms 3D points in the target space to the canonical space. As shown in Fig.~\ref{fig:deformtion}, the deformation field consists of a deformation network (D-Net) and a weighting network (W-Net). 
For each location $x_d$ in the deformed space, we use D-Net to predict its canonical location $x_c$, and then query the canonical tri-Plane feature at $x_c$, which is used to regress the density and color of $x_d$ for neural rendering. In addition to the target expression, we also condition D-Net on source identity to preserve the input identity, which is crucial for endowing it generalization ability.
Inspired by the FLAME mesh model \cite{li2017learning} that assigns skinning weights on mesh vertices for smooth blending, we additionally learn a W-Net to predict offset weights for each location. 

 \subsection{Loss Functions and Training Strategy}
\label{sec:3.3}
\paragraph{General Training Stage.} In the first stage, we train the base model without the compensation branch on a large-scale video dataset in an end-to-end fashion, by sampling source image $I_s$ and target image $I_t$ in the same video clip. During training, we use multiple objectives to ensure faithful reconstruction and vivid reenactment. 

First, we apply a reconstruction loss between the synthesis image $\hat{I}$ and the target $I_t$:
\begin{equation}
\mathcal{L}_{\text{rec}} = {\vert\vert I_t - \hat{I} \vert\vert}_2 + {\rm LPIPS}(I_t,\hat{I}),
\label{eq:recon}
\end{equation}
where ${\rm  LPIPS}(\cdot,\cdot)$ is the perceptual loss ~\cite{zhang2018unreasonable}.
We also use the pixel-wise $L2$ distance to constrain high-level image contents.

We additionally utilize a mouth regularization loss in order to further enhance the mouth region thus derive more accurate facial motion.
Specifically, we crop the mouth region from $I_t$ and $\hat{I}$ using ROI align \cite{he2017mask}, and apply the reconstruction loss on the cropped region, formulated as:
\begin{equation}
\begin{split}
\mathcal{L}_{\text{mouth}} = 
{\vert\vert crop(I_t) - crop(\hat{I}) \vert\vert}_2 + {\rm LPIPS}(crop({I_t}),crop({\hat{I}})),
\label{eq:mouth}
\end{split}
\end{equation}
where $crop(I)$ represents the cropped region of image $I$.

For better identity preservation, we incorporate a face recognition loss between the synthesis image and the target image:
\begin{equation}
    \mathcal{L}_{\text{id}} = 1-\left \langle F(I_t),F(\hat{I}) \right \rangle,
\label{eq:id}
\end{equation}
where $F(\cdot)$ is the pre-trained ArcFace~\cite{deng2019arcface}. $\left \langle \cdot, \cdot \right \rangle$ is cosine distance.  

Finally, we adopt a latent space regularization loss to force the encoder to produce latent codes closer to the average latent code of the pre-trained 3D GAN \cite{eg3d}:
\begin{equation}
\mathcal{L}_{\text{latent}} = {\vert\vert w_s - \overline{w} \vert\vert}_2.
\label{eq:latent}
\end{equation}
It is helpful to regularize the 3D shape of the synthesized volume.  

The total objective function of the general training is defined as:
\begin{equation}
\mathcal{L}_{\text{general}} = 
\lambda_{\text{rec}}\mathcal{L}_{\text{rec}} + 
\lambda_{\text{mouth}}\mathcal{L}_{\text{mouth}} + 
\lambda_{\text{id}}\mathcal{L}_{\text{id}} + 
\lambda_{\text{latent}}\mathcal{L}_{\text{latent}},
\end{equation}
where $\lambda_{\text{rec}},\lambda_{\text{mouth}},\lambda_{\text{id}}$, and $\lambda_{\text{latent}}$ are trade-off hyperparameters. They are set as $1,0.5,0.1$, and $0.01$ respectively.

\paragraph{Teeth Refinement} 
After the first training stage, our model can produce nearly satisfying reconstruction and animation results except for the synthetic teeth, which are of great influence on the visual quality yet ignored by most other approaches. 
It is challenging to synthesize clear teeth due to the limited resolution and quality of the training videos. To tackle this problem, we exploit a face restoration model GFPGAN \cite{wang2021towards} to provide clear teeth as supervision and continue training the base model for teeth refinement while fixing the deformation field. 
Empirically, we observed that applying GFPGAN on the synthetic image $\hat{I}$ as supervision produce better results than applying it on the target image $I_t$. 
We therefore apply the mouth regularization loss between $\hat{I}$ and $\hat{I}^{*} = {\rm GFPGAN}(\hat{I})$ for teeth refinement:
\begin{equation}
\begin{split}
\mathcal{L}_{\text{teeth}} = {\vert\vert crop(\hat{I}^{*}) - crop(\hat{I}) \vert\vert}_2 + {\rm LPIPS}(crop(\hat{I}^{*}),crop({\hat{I}})),
\label{eq:teeth}
\end{split}
\end{equation}

We also incorporate Eq.~\ref{eq:recon}, Eq.~\ref{eq:id}, and Eq.~\ref{eq:latent} in the teeth refinement stage, and the objective function of this stage is:
\begin{equation}
\mathcal{L}_{\text{refine}} = 
\lambda_{\text{rec}}\mathcal{L}_{\text{rec}} + 
\lambda_{\text{teeth}}\mathcal{L}_{\text{teeth}} + 
\lambda_{\text{id}}\mathcal{L}_{\text{id}} + 
\lambda_{\text{latent}}\mathcal{L}_{\text{latent}},
\end{equation}
where we set $\lambda_{\text{rec}} = 0.5,\lambda_{\text{teeth}} = 1,\lambda_{\text{id}} = 0.1$, and $\lambda_{\text{latent}} = 0.01$.

\paragraph{Training Compensation Network.}
We add a compensation network to supplement fine details for the base model. Specifically, we fix the base model and train the compensation network on face image datasets in a self-supervised way. Given the input source image $I_s$, we adopt loss functions in Eq. ~\ref{eq:recon} and Eq. ~\ref{eq:id} between $I_s$ and the reconstructed image $\hat{I}$. In order to ensure depth and view consistency for the canonical volumes after compensation, we further enforce the output compensation volumes of the compensation network to share similar distribution with the original canonical volumes, using the following regularization term:
\begin{equation}
\mathcal{L}_{\text{depth}} = {\vert\vert \mu V_c - V_m \vert\vert}_2,
\label{eq:depth}
\end{equation}
where $V_m$ and $V_c$ are the compensation volumes and canonical volumes respectively.  $\mu=0.1$ denotes a scale factor. 
The objective function of this stage is:
\begin{equation}
\mathcal{L}_{\text{comp}} = 
\lambda_{\text{rec}}\mathcal{L}_{\text{rec}} + 
\lambda_{\text{id}}\mathcal{L}_{\text{id}} + 
\lambda_{\text{depth}}\mathcal{L}_{\text{depth}},
\end{equation}
where we set $\lambda_{\text{rec}} = 1, \lambda_{\text{id}} = 0.1$ and $\lambda_{\text{depth}} = 0.01$. 

\paragraph{One-shot Fine-tuning.}
When facing challenging cases, we can apply fast adaption on the input source image by fine-tuning the compensation network, which is much faster than fine-tuning the base model using PTI \cite{roich2021pivotal}. Specifically, we optimize the parameters of the compensation network using Eq.~\ref{eq:recon}, Eq.~\ref{eq:mouth}, Eq.~\ref{eq:id} and Eq.~\ref{eq:depth}. The total loss function is:
\begin{equation}
\mathcal{L}_{\text{total}} = 
\lambda_{\text{rec}}\mathcal{L}_{\text{rec}} + 
\lambda_{\text{mouth}}\mathcal{L}_{\text{mouth}} +
\lambda_{\text{id}}\mathcal{L}_{\text{id}} + 
\lambda_{\text{depth}}\mathcal{L}_{\text{depth}},
\end{equation}
where we set $\lambda_{\text{rec}} = 1, \lambda_{\text{mouth}} = 0.5, \lambda_{\text{id}} = 0.1$ and $\lambda_{\text{depth}} = 0.01$. 

\begin{figure*}[!t]
  \centering
  \includegraphics[width=0.8\linewidth]{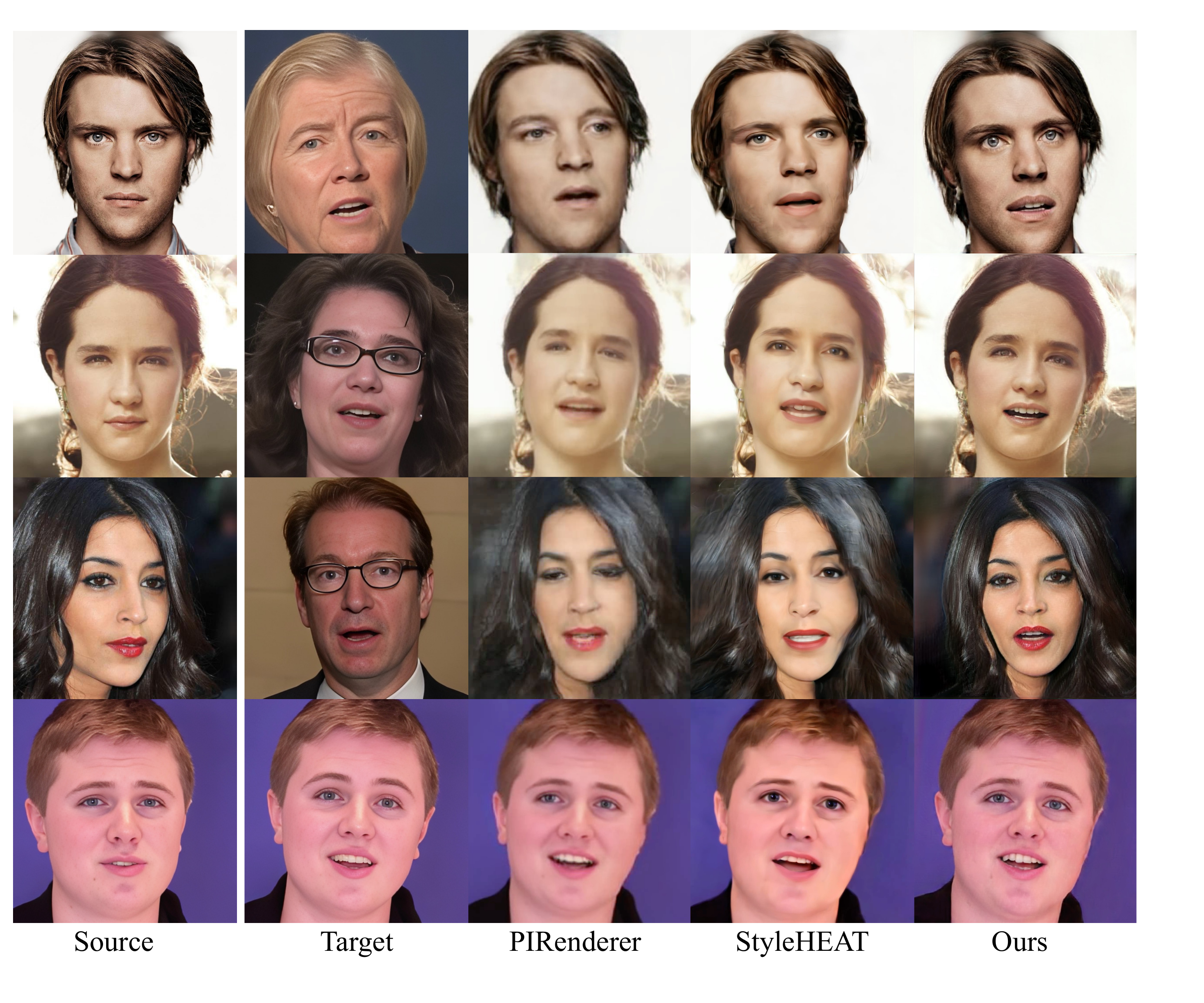}
  \caption{Qualitative comparison with 2D approaches on cross-reenactment (top 3 rows) and self-reenactment (last row). Our approach achieves better reconstruction quality than previous state-of-the-arts.}
  \label{comp_2d_small}
\end{figure*}

\section{Experiments}
\subsection{Implementation Details}
\label{sec:4.1}

\paragraph{Datasets.} 
We train our base model on the CelebV-HQ dataset \cite{zhu2022celebvhq} which contains 35,666 video clips involving 15,653 identities, and train the compensation network on the FFHQ dataset \cite{karras2019style}. We crop and align faces from the videos and extract per-frame 3DMM parameters including identity, expression, and camera parameters for training. The training videos and images are resized into $512 \times 512$.  During inference, we apply camera movement over the reconstructed volumes to obtain head rotation and translation, and can produce continuous head movements without the limitation of the alignment. 

\paragraph{Evaluation Metrics.}
We adopt several metrics to evaluate reconstruction and reenactment quality. The peak signal-to-noise ratio (PSNR),  Structural Similarity (SSIM), and Learned Perceptual Image Patch Similarity (LPIPS) \cite{zhang2018unreasonable} are exploited to evaluate synthetic quality. We also use Frechet In ception Distance (FID) \cite{heusel2017gans} to measure difference between the synthetic and real distributions. 
We calculate the cosine similarity (CSIM) between the source and generated images to evaluate identity preservation. For reenactment quality, we extract 3DMM expression and pose parameters from synthetic and real images to compute their Average Expression Distance (AED) and the Average Pose Distance (APD), following \cite{ren2021pirenderer}.

\paragraph{Training Details.}
Our framework is trained on 8 Nvidia Tesla V100 GPUs in three stages. During training, the ADAM optimizer is adopted with a learning rate of $10^{-4}$. In the general training stage, we train the base model without the compensation network on videos for 250K iterations with a batch size of $16$. In the teeth refinement stage, we fix the deformation field and train the other parts of the base model with the same batch size for 25K iterations. Finally, we fix the base model and train the compensation network on images for 100K iterations with the batch size set to $32$. The training takes about 4 days on 8 V100 GPUs.

\paragraph{Modeling Head Translation.}
Similar to \cite{bergman2022gnarf, wu2022anifacegan, sun2022next3d, tang2022explicitly}, we use camera poses to model head rotation and translation in our implementation. However, the face tracking model \cite{deng2019accurate} used for estimating camera parameters requires face alignment as pre-processing. As a result, the estimated translation parameters are relative translations, and using them directly would cause center-aligned videos, as demonstrated in \cite{bergman2022gnarf, wu2022anifacegan, sun2022next3d, tang2022explicitly}. To describe the absolute head translation, we use facial key-points estimated from the unaligned driving videos to compute pixel offsets between the face center of each driving frame and the source image, and convert them into camera coordinate offsets to rectify camera translation. In this way, the generated face can move freely within a fixed bounding box without the limitation of center align.
\begin{figure*}[t]
  \centering
  \includegraphics[width=1.\linewidth]{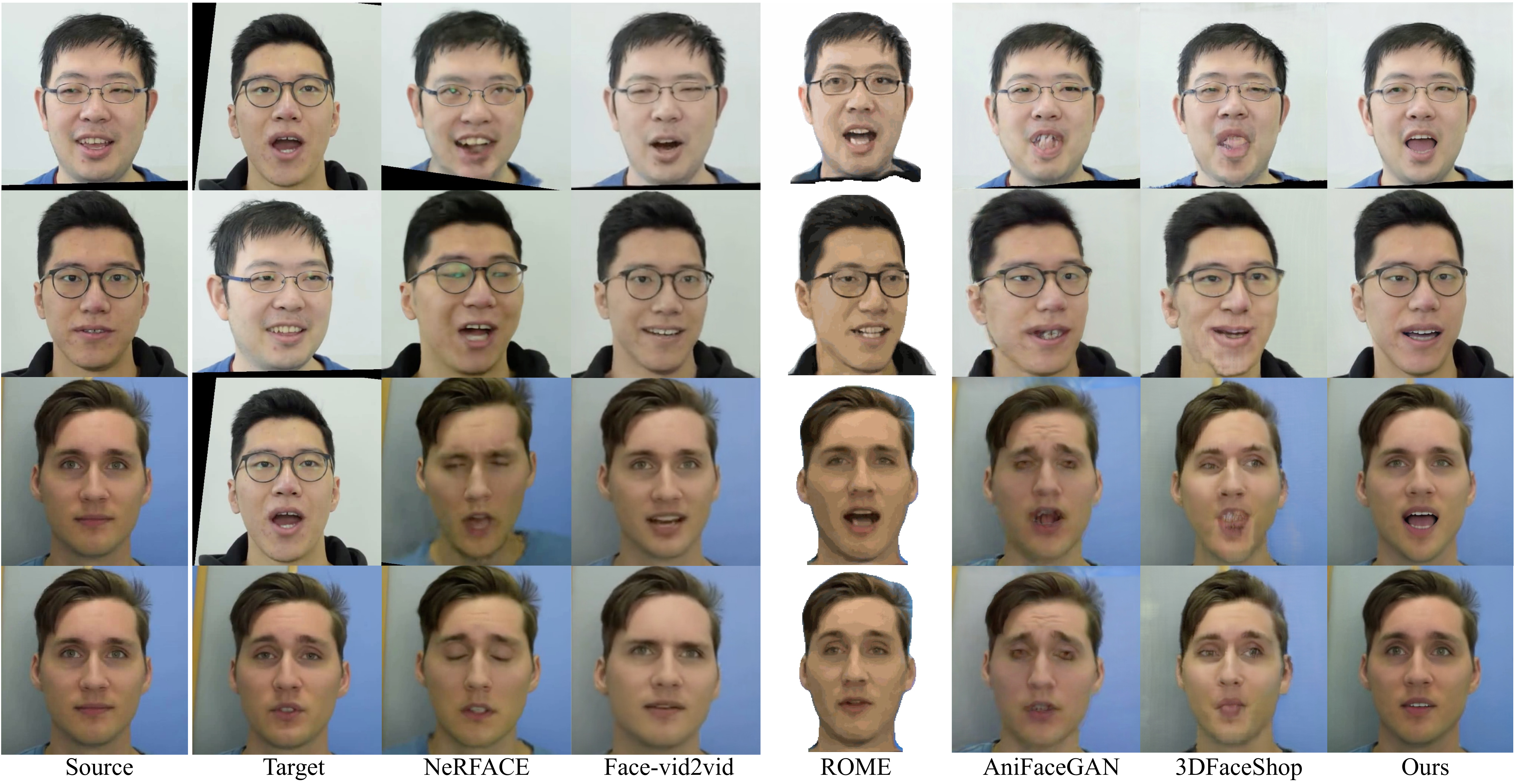}
  \caption{Qualitative comparison with 3D-aware approaches on cross-reenactment
(top 3 rows) and self-reenactment (last row). Our approach synthesizes facial images with higher fidelity and less artifacts comparing with previous state-of-the-arts.}
  \Description{}
  \label{comp_3d}
\end{figure*}
\begin{table}[t]
    \centering
    \small      
    \vspace{-1pt}
    \scalebox{0.8}{
    \begin{tabular}{l|ccccccc}
    \toprule[1pt]
    \textbf{Same-ID} & FID $\downarrow$ & LPIPS $\downarrow$  & PSNR $\uparrow$ & SSIM $\uparrow$  & CSIM $\uparrow$ & AED $\downarrow$ & APD $\downarrow$ \\
    \hline
   PIRenderer~\cite{ren2021pirenderer}  & $27.14$ & $0.2252$ & $30.96$ & $0.6028$ & $0.7797$ & $\textbf{0.1073}$ & $0.01459$\\
   StyleHEAT~\cite{yin2022styleheat} & $18.02$ & $0.1729$ & $31.21$ & $0.6019$ & $0.7475$ & $0.1151$ & $0.01664$\\
   Ours & $\textbf{16.94}$ & $\textbf{0.1481}$ & $\textbf{31.78}$ & $\textbf{0.6175}$ & $\textbf{0.8031}$ & $0.1091$ & $\textbf{0.01142}$\\
    \midrule[1pt]
    \textbf{Cross-ID} & FID $\downarrow$ & $-$  & $-$ & $-$  & CSIM $\uparrow$ & AED $\downarrow$ & APD $\downarrow$ \\
    \hline
 PIRenderer~\cite{ren2021pirenderer}  & $108.56$ & $-$  & $-$ & $-$& $0.4812$ & $\textbf{0.2554}$ & $0.02962$  \\ 

 StyleHEAT~\cite{yin2022styleheat} & $91.28$ & $-$  & $-$ & $-$& $0.4890$ & $0.2630$ & $0.03484$  \\ 

Ours & $\textbf{84.47}$ & $-$  & $-$ & $-$& $\textbf{0.5397}$ & $0.2581$ & $\textbf{0.01633}$  \\ 

    \bottomrule[1pt]
    \end{tabular}
}
\vspace{2pt}
\caption{Quantitative comparison with 2D approaches on face reenactment. The metrics indicate that our approach achieves the best reconstruction quality, comparable expression accuracy and the best pose accuracy compared with 2D approaches.}
\label{metrics_2d}
\end{table}
\begin{table}[t]
    \centering
    \small      
    \vspace{-1pt}
    \scalebox{0.8}{
    \begin{tabular}{l|ccccccc}
    \toprule[1pt]
    \textbf{Same-ID} & FID $\downarrow$ & LPIPS $\downarrow$  & PSNR $\uparrow$ & SSIM $\uparrow$  & CSIM $\uparrow$ & AED $\downarrow$ & APD $\downarrow$ \\
    \hline
   NeRFACE~\cite{gafni2021dynamic}  & $\textbf{12.58}$ & $\textbf{0.0925}$ & $30.87$ & $\textbf{0.6637}$ & $0.7846$ & $0.1103$ & $\textbf{0.01071}$\\
   Face-vid2vid~\cite{wang2021one} &$20.69$ & $0.2035$ & $30.98$ &$0.6191$ &$0.7912$ &$\textbf{0.0995}$ &$0.01250$ \\
   ROME~\cite{khakhulin2022realistic} &$25.83$ & $0.2314$ & $30.85$ &$0.6007$ &$0.7206$ &$0.1224$ &$0.01223$ \\
  AniFaceGAN~\cite{wu2022anifacegan} &$24.39$ & $0.2105$ & $29.86$ &$0.6033$ &$0.7754$ &$0.1480$ &$0.01288$ \\
  3DFaceShop~\cite{tang2022explicitly} &$22.75$ & $0.2154$ & $30.07$ &$0.6095$ &$0.7516$ &$0.1391$ &$0.01167$ \\
   Ours & $16.41$ & $0.1377$ & $\textbf{31.09}$ & $0.6175$ & $\textbf{0.7953}$ & $0.1195$ & $0.01128$\\

    \midrule[1pt]
    \textbf{Cross-ID} & FID $\downarrow$ & $-$  & $-$ & $-$  & CSIM $\uparrow$ & AED $\downarrow$ & APD $\downarrow$ \\
    \hline
 NeRFACE~\cite{gafni2021dynamic}  & $157.38$ & $-$  & $-$ & $-$& $0.3504$ & $0.2554$ & $0.02391$  \\ 
  Face-vid2vid~\cite{wang2021one}  & $92.41$ & $-$  & $-$ & $-$& $0.5024$ & $\textbf{0.2369}$ & $ 0.02541$  \\
 ROME~\cite{khakhulin2022realistic}  & $95.07$ & $-$  & $-$ & $-$& $0.4693$ & $0.2670$ & $0.02011$  \\
  AniFaceGAN~\cite{wu2022anifacegan}  & $93.47$ & $-$  & $-$ & $-$& $0.4944$ & $0.2721$ & $ 0.02150$  \\
  3DFaceShop~\cite{tang2022explicitly}  & $92.53$ & $-$  & $-$ & $-$& $0.5172$ & $0.2855$ & $ 0.02084$  \\
Ours & $\textbf{88.61}$ & $-$  & $-$ & $-$& $\textbf{0.5524}$ & $0.2438$ & $\textbf{0.01872}$  \\ 

    \bottomrule[1pt]
    \end{tabular}
}
\vspace{2pt}
\caption{Quantitative comparison with 3D-aware approaches. In self-reenactment, our one-shot approach achieves comparable performance against NeRFACE that is trained on 1K frames, and surpasses the other approaches. In cross-reenactment, our approach outperforms the other approaches across most of the metrics, and achieves the second-highest expression accuracy.}
\label{metrics_3d}
\end{table}
\subsection{Comparisons}
\label{sec:4.2} 
\paragraph{Comparison with 2D Reenactment Approaches.}
We first evaluate our approach in comparison with two state-of-the-art 3DMM-based 2D talking face generation approaches: PIRenderer \cite{ren2021pirenderer} and StyleHEAT \cite{yin2022styleheat}. Both of them supports pose control via 3DMM parameters and video-driven face reenactment. 

For face reenactment evaluation, we conduct two types of reenactment tasks, \textit{i.e.,} self-reenactment and cross-reenactment. For self-reenactment, the identity of source image is the same with the driving frames; for cross-reenactment, the source image and driving frames come from two different identities. The latter setting is much more challenging because of the facial feature gap between the source and driving faces. Following \cite{yin2022styleheat}, we use 20 video clips with a total of 10K frames from HDTF dataset \cite{zhang2021flow} for self-reenactment evaluation. For cross-reenactment evaluation, we use the first 1,000 images from the CelebA-HQ dataset \cite{karras2018progressive} as source images and the 20 HDTF videos as driving videos. Similar as \cite{yin2022styleheat}, we perform one-shot fine-tuning on the source image to achieve better visual quality.

Fig.~\ref{comp_2d_small} shows the qualitative reenactment results of our approach and other state-of-the-arts, where our approach achieves better reconstruction quality in terms of identity preservation and detail textures. When dealing with side faces, feature warping-based 2D approaches fails to inference reasonable frontal faces and suffers from severe artifacts, while our approach successfully reconstruct accurate and reasonable frontal face. Table~\ref{metrics_2d} shows the quantitative evaluation results, which demonstrate that our approach achieves better reconstruction results. For facial motion modeling, the average expression distance (AED) indicates that our 3D approach yields competitive animation results compared with 2D PIRenderer ~\cite{ren2021pirenderer}. For head rotation modeling, our approach surpasses the two baselines with a large margin in terms of the average pose distance (APD), because the head poses are directly controlled by the external camera parameters in our implementation. 
\paragraph{Comparison with 3D-aware Reenactment Approaches.}
We further compare with several 3D-aware approaches: NeRFACE \cite{gafni2021dynamic}, Face-vid2vid \cite{wang2021one}, ROME \cite{khakhulin2022realistic}, AnifaceGAN \cite{wu2022anifacegan} and 3DFaceShop \cite{tang2022explicitly}. Specifically, NeRFACE is a multi-shot NeRF avatar reconstruction approach. AnifaceGAN and 3DFaceShop are unconditional 3D GANs, we adopt PTI \cite{roich2021pivotal} to apply them on real images. We conduct self- and cross-reenactment experiments on the NeRFACE dataset \cite{gafni2021dynamic}, where we use the first 1K frames of each video to train NeRFACE, while keep the one-shot setting of other approaches.

Fig.~\ref{comp_3d} shows the qualitative results. 
For self-reenactment, NeRFACE generates several expressions inconsistent with the driving image; for cross-reenactment, NeRFACE suffers from severe artifacts and produces inconsistent expressions. What's more, Face-vid2vid produces inaccurate poses because it doesn't model the 3D geometry.
ROME suffers from oversmoothed appearance due to its mesh representation.
AnifaceGAN and 3DFaceShop fail to keep fidelity under unseen views, and suffer from inaccurate motion and teeth artifacts. In contrast, our model ensures fine-grained motion control and high-fidelity across views.
The quantitative results are listed in Table.~\ref{metrics_3d}. For self-reenactment evaluation, our one-shot approach achieves comparable performance against NeRFACE that is trained on 1K frames, and surpasses the other approaches. More importantly, In the more challenging and applied cross-reenactment task, 
our approach outperforms the other competitors across most of the metrics, and achieves the second-highest expression accuracy. 

\subsection{Ablation Study}
\label{sec:4.3}
\begin{figure}[t]
  \centering
  \includegraphics[width=\linewidth]{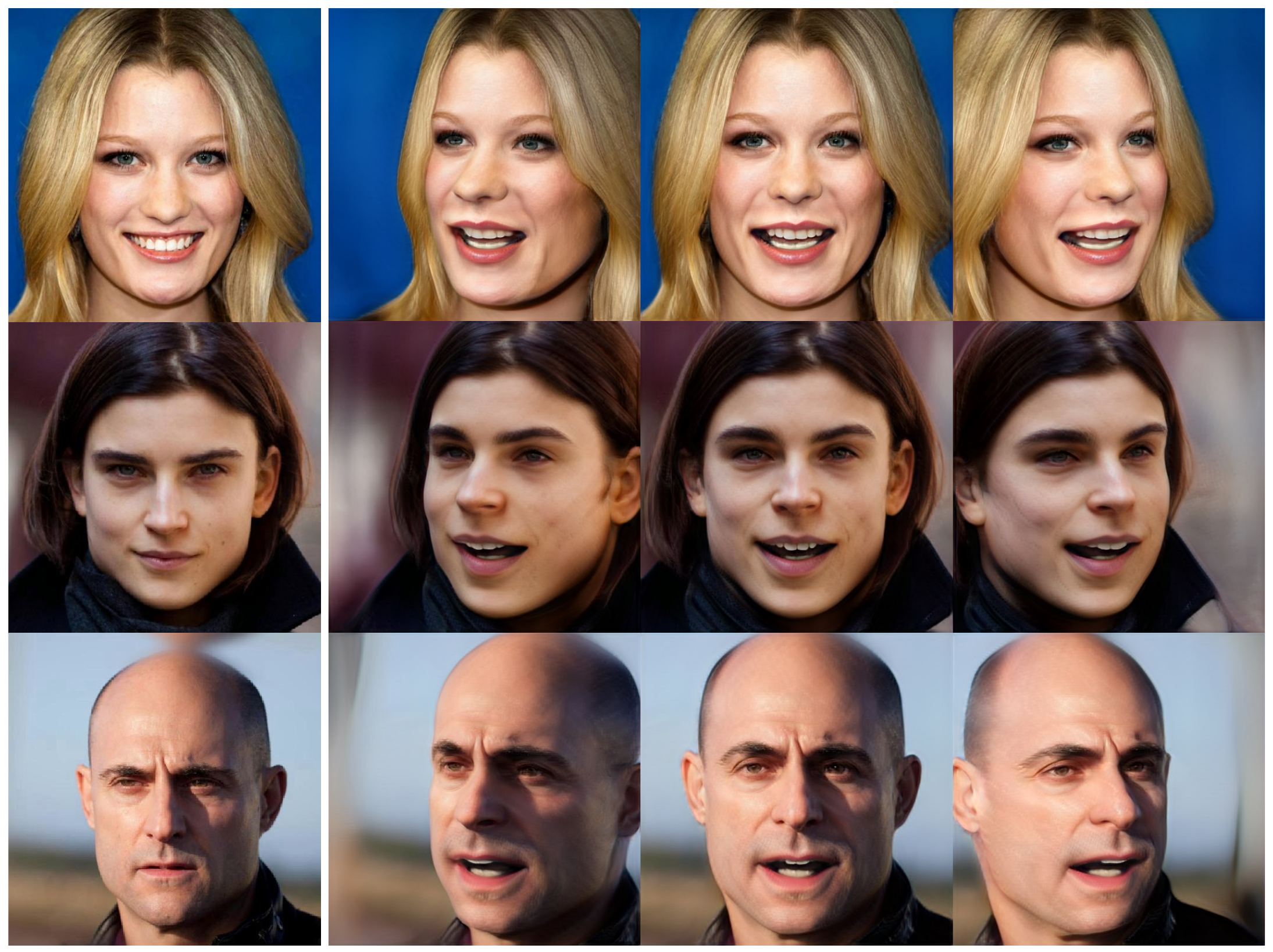}
  \caption{Visualization of canonical space. The synthetic images are directly rendered from canonical volumes without deformation, sharing the same aligned expression}
  \Description{}
  \label{canonical}
\end{figure}


\begin{figure}[h]
  \centering
  \includegraphics[width=\linewidth]{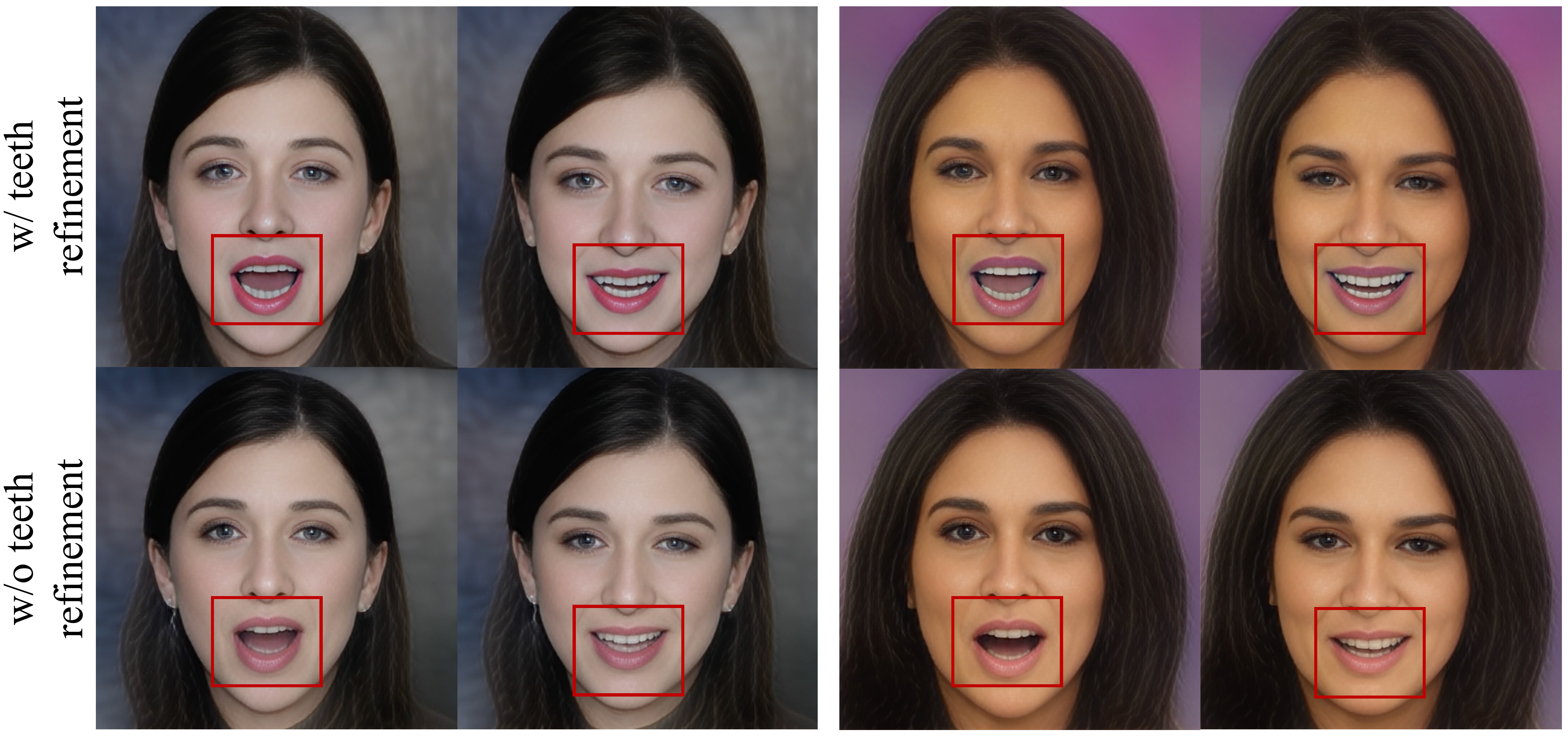}
  \caption{Qualitative evaluation of teeth refinement. It helps to synthesize clearer teeth with less artifacts.}
  \label{teeth}
\end{figure}
\paragraph{Visualization of Canonical Space.}
Given the source image, our model produces neural volumes in the canonical space instead of preserving the original expression of source image. Given the target expressions, the back-ward deformation is applied by querying volumes in the canonical space, so that the canonical volumes are deformed into target expression. In our implementation, the canonical space is naturally learned by jointly training the whole framework on videos in an end-to-end manner, without explicit inner supervision. Fig.~\ref{canonical} shows examples of the canonical space, the output images are directly rendered from canonical volumes without deformation, sharing the same aligned expression.

\begin{figure}[h]
  \centering
  \includegraphics[width=\linewidth]{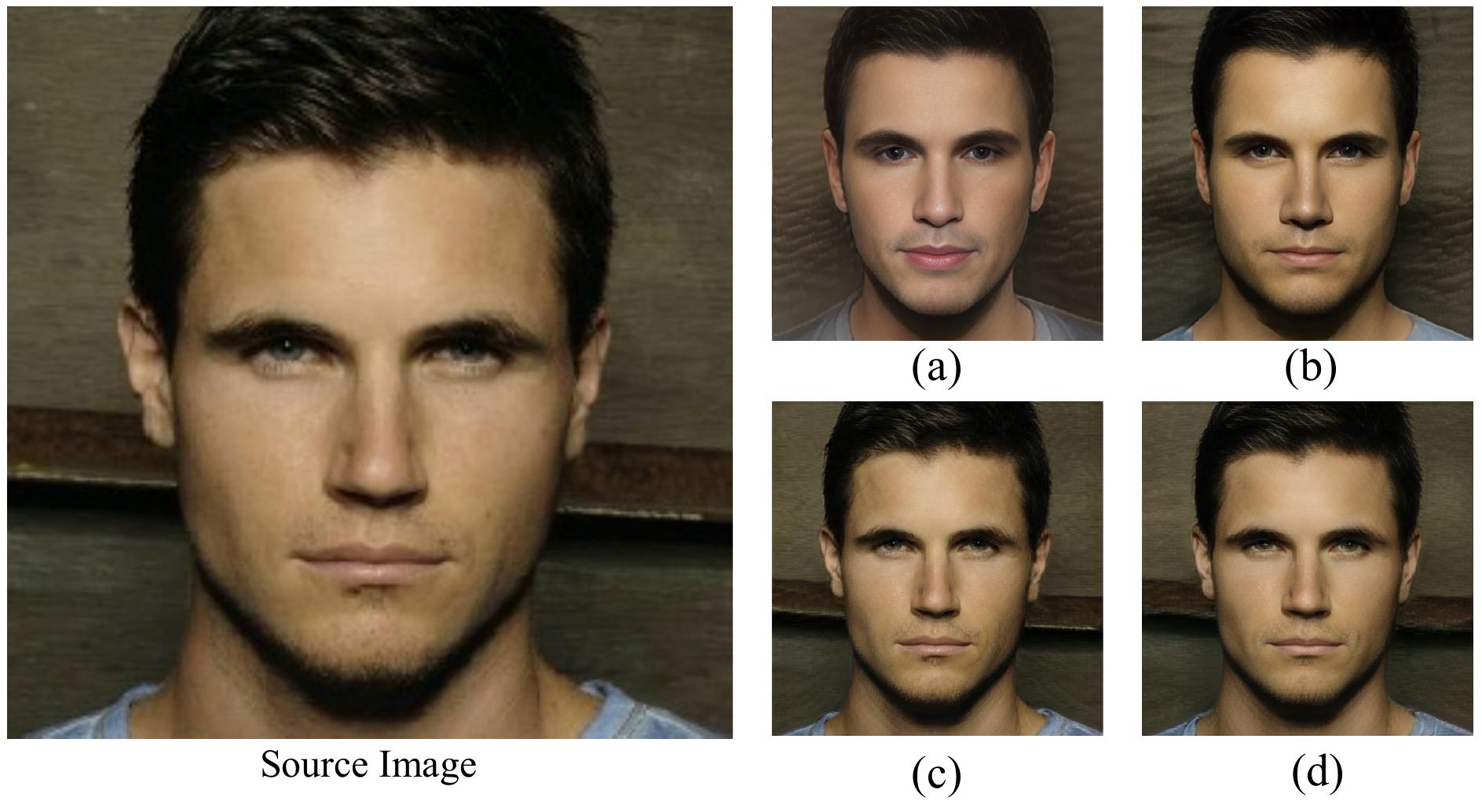}
  \caption{Qualitative evaluation of the compensation network. (a). w/o compensation network, (b). w/ compensation network, (c). optimize generator using PTI \cite{roich2021pivotal}, (d). optimize compensation network.}
  \Description{}
  \label{ablation_compensation}
\end{figure}
\paragraph{Evaluation of Compensation Network.}
We design a compensation network to supplement identity and texture information for the canonical volumes. 
As shown in Fig.~\ref{ablation_compensation} (a) and (b), given the source image, our base model yields correct 3D shape and expression yet fails to preserve the source identity. With the compensation network, the reconstruction quality is greatly improved. 
Our approach also supports one-shot fine-tuning to achieve more accurate reconstruction. As shown in Fig.~\ref{ablation_compensation} (c) and (d), 
fine-tuning the whole generator using PTI\cite{roich2021pivotal} produces photo-realistic reconstruction results, yet the training costs about \textbf{20} minutes on \textbf{4} V100 GPUs. With the compensation network, we can achieve comparable reconstruction results in a more time- and memory- efficient way, simply by optimizing the compensation network, which only takes about \textbf{10} minutes on a \textbf{singe} V100 GPU. 
We conducted self-reenactment experiments in Sec. ~\ref{sec:4.2}, with the four different settings. Table. ~\ref{table_compensate} shows the quantitative evaluation, which further demonstrate the effectiveness of the compensation network.

\paragraph{Effectiveness of Teeth Refinement.}
We incorporated a teeth refinement training stage to enable the base model to synthesize clearer teeth. Fig.~\ref{teeth} shows the faces generated before and after the teeth refinement stage. After teeth refinement, the synthetic teeth are clearer and show less artifacts, while without it the synthetic teeth tend to be blurry. We conduct self-reenactment experiment on the HDTF \cite{zhang2021flow} dataset, and crop the mouth region of the synthetic images to compute metrics, the results are listed in Table.\ref{table_teeth}. Through the teeth refinement stage, the quality of the reconstructed teeth improved.
\paragraph{Evaluation of the Deformation Field.}
We model facial dynamics using a deformation field. Typically, we learn a weighting network to increase motion accuracy, and use the 3DMM identity parameter as the condition for both the deformation network and the weighting network to preserve the source identity. We conduct self-reenactment experiment on the HDTF dataset to evaluate the designs. Specifically, we use the average expression distance (AED) to evaluate the effectiveness of the weighting network, and use CSIM to evaluate the effectiveness of identity condition. The results are listed in Table.\ref{table_deformation}, which demonstrate the effectiveness of the proposed designs.
\begin{table}[t]
    \centering
    \small   
    \scalebox{0.85}{
    \begin{tabular}{l|ccccc}
    \toprule[1pt]
    \textbf{Settings} & FID $\downarrow$ & LPIPS $\downarrow$  & PSNR $\uparrow$ & SSIM $\uparrow$  & CSIM $\uparrow$  \\
    \hline
    w/o Compensation & $36.45$ & $0.2429$ & $29.75$ & $0.5836$ & $0.6024$ \\
   w/ Compensation & $25.63$ & $0.2240$ & $31.05$ & $0.6007$ & $0.7219$ \\
    Optimize compensation network & $18.75$ & $0.1697$ & $\textbf{31.94}$ & $0.6042$ & $0.7981$ \\
   Optimize generator & $\textbf{16.94}$ & $\textbf{0.1481}$ & 31.78 & $\textbf{0.6175}$ & $\textbf{0.8031}$  \\
    \bottomrule[1pt]
    \end{tabular}
}
\vspace{2pt}
\caption{Quantitative evaluation of the compensation network. The reconstruction quality is greatly improved with compensation. What's more, optimizing the compensation network results in a speed increase of eight times while achieving performance gains comparable to optimizing the entire generator.}
\label{table_compensate}
\end{table}

\begin{table}[t]
    \centering
    \small
    \vspace{-1pt}
    \scalebox{1}{
    \begin{tabular}{l|ccccc}
    \toprule[1pt]
    \textbf{Settings} & FID $\downarrow$ & LPIPS $\downarrow$ & SSIM $\uparrow$   \\
    \hline
    w/o teeth refinement & $28.45$ & $0.2678$ & $0.3297$  \\
   w/ teeth refinement & $\textbf{25.31}$ & $\textbf{0.2215}$ & $\textbf{0.3478}$   \\
    \bottomrule[1pt]
    \end{tabular}
}
\caption{Evaluation of teeth refinement. The quality of the synthetic teeth improved through the teeth refinement stage,.}
\label{table_teeth}
\end{table}
\begin{table}[t]
    \centering
    \small
    \vspace{-1pt}
    \scalebox{1}{
    \begin{tabular}{c|cc}
    \toprule[1pt]
    \textbf{Weighting network} & w/ & w/o \\
    \hline
   AED $\downarrow$ & $\textbf{0.1091}$ & $0.1143$ \\
    \midrule[1pt]
    \textbf{Identity condition} & w/ & w/o  \\
    \hline
    CSIM $\uparrow$ & $\textbf{0.8031}$ & 0.7210   \\ 

    \bottomrule[1pt]
    \end{tabular}
}
\caption{Evaluation of weighting network and identity condition in the deformation field.}       
\label{table_deformation}
\end{table}

\section{limitations and ethical issues}


In our implementation, the head poses are modeled as camera poses, thus the background rotates as head rotates. We will explore the background modeling in future works.

Since our framework is capable of reconstructing high-fidelity facial avatar using only a single image, it may pose the risk of nefarious use such as deep-fakes. We are keenly aware of the potential for abuse of our approach, and we will explore the implementation of robust video watermarks for the synthesized videos, as well as develop tools to verify the authenticity.


\bibliographystyle{ACM-Reference-Format}
\bibliography{refs}

\clearpage

\end{document}